\documentclass[]{interact}
\usepackage{xcolor}
\usepackage{epstopdf}% To incorporate .eps illustrations using PDFLaTeX, etc.
\usepackage[caption=false]{subfig}% Support for small, `sub' figures and tables
\usepackage{caption}
\usepackage{hyperref}
\usepackage{fancyhdr}

\usepackage[round,sort&compress]{natbib}
\setcitestyle{aysep={}}

\bibpunct[, ]{(}{)}{;}{a}{}{,}% Citation support using natbib.sty
\theoremstyle{plain}% Theorem-like structures provided by amsthm.sty

\theoremstyle{definition}

\theoremstyle{remark}

\fancypagestyle{FirstPage}{
\lfoot{
\footnotesize{This is an Accepted Manuscript version of the following article, accepted for publication in Computer Methods in Biomechanics and Biomedical Engineering: Imaging \& Visualization \citep{sharan_doi:10.1080/21681163.2022.2159535}. It is deposited under the terms of the Creative Commons Attribution-NonCommercial License (http://creativecommons.org/licenses/by-nc/4.0/), which permits non-commercial re-use, distribution, and reproduction in any medium, provided the original work is properly cited.}
}
}

\begin{document}

\title{mvHOTA: A multi-view higher order tracking accuracy metric to measure temporal and spatial associations in multi-point tracking}

\author{
\name{Lalith Sharan\textsuperscript{a,b 0000-0003-0835-042X}\thanks{CONTACT Lalith Sharan Email: lalithnag.sharangururaj@med.uni-heidelberg.de}, % Orcid ID included
Halvar Kelm\textsuperscript{a},
Gabriele Romano\textsuperscript{a 0000-0002-6843-4492},
Matthias Karck\textsuperscript{a},
Raffaele De Simone\textsuperscript{a},
Sandy Engelhardt \textsuperscript{a,b 0000-0001-8816-7654}} % Orcid ID included
\affil{\textsuperscript{a}Department of Cardiac Surgery, Heidelberg University Hospital, Heidelberg, Germany; \textsuperscript{b}DZHK (German Centre for Cardiovascular Research), partner site Heidelberg/Mannheim, Germany}
}

\maketitle

\begin{abstract}
Multi-point tracking is a challenging task that involves detecting points in the scene and tracking them across time. Here, metrics from Multi-object tracking (MOT) methods are shown to perform better than frame-based F-measures. The recently proposed HOTA metric, used for benchmarks such as the KITTI dataset, better evaluates the performance over metrics like MOTA, DetA, and IDF1. While HOTA takes into account temporal associations, it does not provide a tailored means to analyse the spatial associations of a dataset in a multi-camera setup. Moreover, there are differences in evaluating the detection task for points vs. objects (point distances vs. bounding box overlap). Therefore, we propose a multi-view higher-order tracking metric mvHOTA , to determine the accuracy of multi-point (multi-instance and multi-class) tracking methods while taking into account temporal and spatial associations. We demonstrate its use in evaluating the tracking performance on an endoscopic point detection dataset from a previously organised surgical data science challenge. Furthermore, we compare with other adjusted MOT metrics for this use-case, discuss the properties of mvHOTA, and show how the proposed multi-view Association and the Occlusion index (OI) facilitate analysis of methods with respect to handling of occlusions. 
The code is available at \url{https://github.com/Cardio-AI/mvhota}.
\end{abstract}


% -------------- INTRODUCTION ---------------------------------------------
\section{Introduction}
\label{sec:Introduction}

\thispagestyle{FirstPage}

Recent seminal works in our field by the MICCAI Special Interest Group Biomedical Image Analysis Challenges (SIG-BIAC) (\citep{reinke_common_2021,maier-hein_why_2018}) have emphasized the considerable impact of evaluation metrics in benchmarking and ranking the performance of different machine learning methods. The authors recommend usage of orthogonal performance criteria and careful interpretation of the results with respect to the task definition. However, little emphasis so far has been given to metrics that  consider a temporal correlation of data. Given that this is a key aspect in Surgical Data Science (SDS), where image modalities such as ultrasound, fluoroscopy or endoscopy are traditionally temporally resolved to provide real-time guidance, we found that it is interesting to our field to focus on this in more detail.
Key use-cases in the domain of SDS require the tracking of objects over time, e.g., surgical instruments, catheters and other equipment, anatomical landmarks, pathological structures, etc. Specific applications are mitral valve leaflet tracking from echocardiography \citep{chandra_mitral_2020}, or real-time tracking of aortic valve landmarks from fluoroscopy \citep{Karar2010RealTimeTO}.
Furthermore, many detection-based methods exist, such as endoscopic surgical action triplet detection \citep{zia_endoscopic_2022}, 
polyp detection \citep{brandao_towards_2018} and artefact detection \citep{yin_endoscopy_2022}. However, in these methods, frame-level metrics such as the $F$-score, \textit{precision}, and \textit{recall} are computed for the frames of the sequence. 
Besides, one can differentiate between single- and  multi-view settings, where the same scene or object is captured from different angles. This is typically the case in stereo-endoscopy, where spatial associations need to be identified between multiple views. There is hitherto little emphasis given to the robustness of both temporal and spatial consistency of the detection-based results. 
In this work, we extend a Multi-object tracking (MOT) benchmark metric for the point tracking use-case to account for temporal and spatial associations, and analyse the results on an endoscopic dataset.

In particular, the use-case of point tracking forms a special case of a Multi-object tracking (MOT) task, where several points should be identified. They can either represent multiple instances of the same class (e.g., the same type of cell), or represent multiple classes (e.g., the tip of different surgical instruments). We refer to it as a \textit{multi-point tracking} problem.  
In endoscopy, multi-point tracking is relevant for applications such as respiratory motion estimation \citep{silverstein_comparative_2018}, and surgical suture detection \citep{sharan_point_2021}, besides having more general applications in the computer vision domain spanning motion tracking in traffic scenes \citep{Geiger2012CVPR}, and pose estimation \citep{8765346}.
In the context of mitral valve repair, a minimally invasive surgery of the heart valve, the detection and tracking of entry and exit points of sutures from an endoscopic image is a useful task for understanding a surgical scene. It enables a potential comparison between different skill levels, analysis of suture configurations, and can function as markers for augmented reality visualisations \citep{10.1007/978-3-319-10437-9_14}. However, multi-point tracking is a challenging task due to objects moving in and out of the scene during a surgery, leading to occlusions in the temporal domain. 
Furthermore in a multi-view setting, the same point can be contained in all perspectives or might be occluded in some.
In reality a point may be occluded from one frame to another, may move out of the scene and reappear, or be visible in only one of the views. This is crucial in applications that involve keypoint matching or point cloud generation. While multiple methods address one or more of these issues, the proper evaluation of these methods is difficult in more complex scenarios, for all involved subtasks of finding temporal and spatial associations, \textit{and} detection of the points themselves. Hence, developing a single metric which treats all of these subtasks as equally important is crucial.

\begin{figure}[b]
\includegraphics[width=\textwidth]{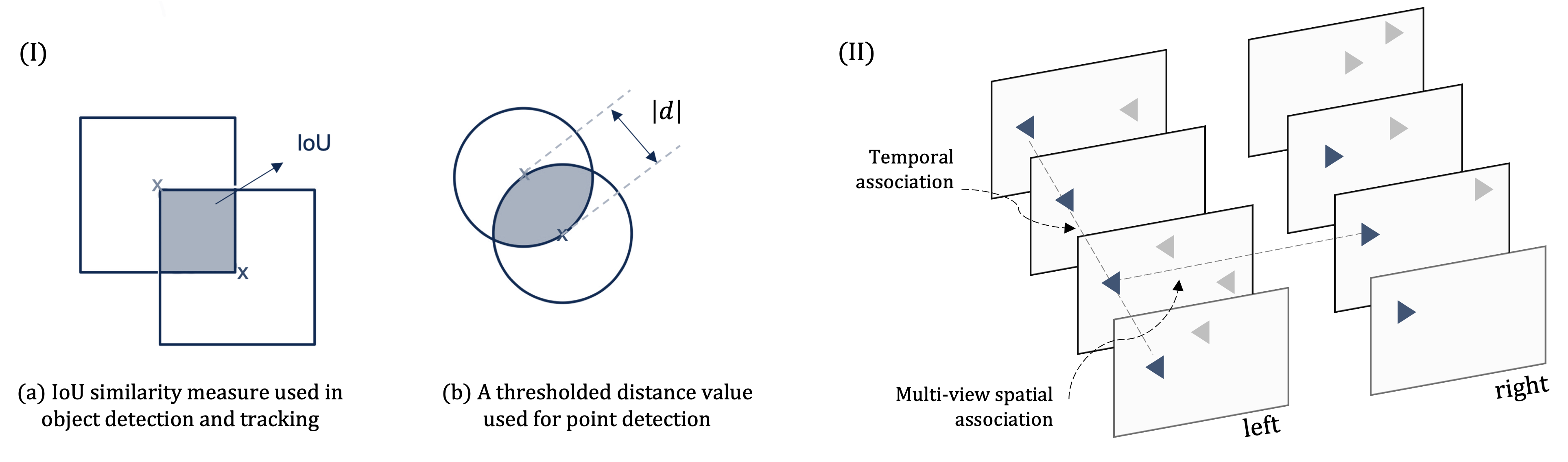}
\caption{(Ia) The Intersection over Union (spatial \textit{Jaccard} formulation) used in object detection vs. (Ib) thresholded radius used in point detection. (II) A sample case that shows the detection, temporal and spatial associations.} 
\label{fig:overview}
\end{figure}

Although multi-point tracking is a closely related task to, and can be considered a special case of a multi-object tracking task \citep{Geiger2012CVPR,dendorfer_cvpr19_2019}, important differences exist in evaluating the intrinsic detection task. 
In MOT, bounding boxes are predicted for each object, where an \textit{Intersection over Union} (IoU) is typically used to compute the overlap of a source and target bounding box (see Fig. \ref{fig:overview}(Ia)). The IoU has the value $0$ when there is no overlap, and $1$ when there is complete overlap. Then, a matching algorithm is employed to match the ground-truth labels and predictions. The matches obtained through this algorithm are then filtered with a threshold $\alpha$, where the label-prediction pairs with an overlap greater than $\alpha$ are considered a successful match. 
The algorithm maximises the global similarity and the number of true positive detections \citep{luiten_hota_2021}. 
In contrast in point detection, the minimum distance between two points is $0$, but the maximum distance is theoretically unbounded. Practically, the maximum distance between two points in an image is bounded by the image diagonal, e.g., as shown by \cite{koehler_10.1007/978-3-658-36932-3_43}. Moreover, unlike bounding boxes, comparison in the case of point detection is equivalent to considering a radius centered around the point (see Fig. \ref{fig:overview}(Ib)). 

The previously proposed \textit{Higher Order Tracking Accuracy} (HOTA) metric \citep{luiten_hota_2021} equally weights the detection and temporal association, and is used by the KITTI \citep{Geiger2012CVPR} and MOT \citep{dendorfer_cvpr19_2019} benchmarks to evaluate submitted methods. 
In this work, we extend the HOTA metric \citep{luiten_hota_2021} to evaluate the multi-view higher order tracking accuracy (\textit{mvHOTA}) by incorporating both temporal and spatial associations (see Fig. \ref{fig:overview}(II)). This is realized by computing a \textit{multi-view Association (mvAssc)} along with detection and temporal association. While previous work \citep{luiten_hota_2021} presents a trivial extension to a multi-view setup, here we propose a metric that can be used to analyse the handling of both temporal and spatial occlusions. We demonstrate how this method can be used to evaluate and analyse a point tracking method on a surgical stereo-endoscopic dataset of mitral valve repair \citep{engelhardt_deep_2021}.  
In addition, we introduce the computation of a \textit{Occlusion Index} for multi-camera setups to indicate the amount of occlusions in both the temporal and spatial domains of a dataset. 
Furthermore, we present an analysis of the metric properties, and a comparison to other MOT metrics for the point tracking use case.

% -------------- RELATED WORK ------------------------------------------------
\section{Related Work}
\label{sec:related_work}
Currently, multiple metrics exist to evaluate a tracking task from multi-object tracking (MOT) applications \citep{bernardin_evaluating_2008,ristani_10.1007/978-3-319-48881-3_2,vedaldi_tao_2020}. The CLEAR-MOT \citep{bernardin_evaluating_2008} metrics were introduced as a standard to evaluate a range of tracking methods, and has been used for years as the standard to benchmark a range of MOT methods. Primarily, the \textit{Multi Object Tracking Accuracy (MOTA)} and \textit{Multi Object Tracking Precision (MOTP)} are used to track the association of detection over time.
The $IDF1$ proposed by \cite{ristani_10.1007/978-3-319-48881-3_2} is another metric initially proposed for use in multi-target multi-camera tracking systems and later also adopted for MOT evaluation. 
Track-mAP is commonly used in benchmarks such as Image-Net  \citep{russakovsky_imagenet_2015}, and TAO \citep{vedaldi_tao_2020}, which requires the use of confidence scores along with the tracker predictions. 
The recently proposed \textit{Higher Order Tracking Accuracy} (HOTA) introduced by \cite{luiten_hota_2021} has replaced many of the multi-object tracking benchmarks, for example in KITTI MOT \citep{Geiger2012CVPR} and the MOTS challenge \citep{dendorfer_cvpr19_2019}. Besides proposing a single unifying metric to measure detection and association, the work analyses the drawbacks of the previously used MOTA and IDF1 metrics in terms of monotonicity and error type differentiability, as explained by \cite{leichter_10.1109/TPAMI.2013.70}.

% -------------- METHODS ------------------------------------------------
\section{Methods}
\label{sec:methods}

The goal is to evaluate point detection methods not only on a per-frame basis but across temporal sequences and a multi-camera setup, taking into account both temporal and spatial associations. This means, it is not enough to quantify the detection performance on a frame level (Section \ref{subsec:detection}), but also define the associated detection in the temporal domain, which we call temporal associations (Section \ref{subsec:temp_assc}), and the detection across multiple views, which we call spatial associations (Section \ref{subsec:mv_assc}). In other words, the final metric incorporates how well a point is detected, and additionally tracked over time and space. 
While the previously proposed \textit{Higher order tracking accuracy (HOTA)} \citep{luiten_hota_2021} formulates the detection and temporal association for a MOT task, here we propose a metric to also include the multi-view spatial association.
The use of a higher-order metric enables us to benchmark and compare methods with an aggregate, while at the same time decompose the metric into different source of tracking errors \citep{leichter_10.1109/TPAMI.2013.70} that can be attributed to per-frame detection, temporal and spatial associations.

\subsection{Pre-requisites}
\label{subsec:pre_req}
The computation of this metric requires that we are able to match a predicted point ($p_{pred}$) to a ground-truth point ($p_{gt}$) in a frame, and additionally track this particular point over time and the different camera views. This means, a $p_{gt}$ needs a unique ID with which it can be identified over different time steps and across the different camera views.
These labels are typically assigned during the annotation process. Similarly, a predicted point $p_{pred}$ requires a consistent ID across a temporal sequence and camera views. This can be directly obtained from the predictions of a multi-view tracking model, or alternatively the predictions can be matched using a temporal matching algorithm (cf. Section \ref{subsec:id}).
The spatial associations across the different camera views can be inferred by matching a $p_{pred}$ with a $p_{gt}$ in a view, since we have the spatial matches for the $p_{gt}$ points between the views. With this preparation, we are now ready to formulate the detection, temporal association, and spatial association, and combine it into the \textit{multi-view Higher Order Tracking Accuracy (mvHOTA)}.

\subsection{Detection}
\label{subsec:detection}
We first define \textit{True Positive (TP)}, \textit{False Positive (FP)} and \textit{False Negative (FN)} detections for an image. The points that lie within the detection threshold radius of $\alpha$ are classified as TP. The $\{p_{gt}\}$ without a matched prediction are the FN, and the set of predictions $\{p_{pred}\}$ without a matched ground-truth point are the FPs.

For every frame, the set of ground-truth points $\{p_{gt}\}$, and the set of predicted points $\{p_{pred}\}$ are matched using the Hungarian matching algorithm \citep{kuhn_hungarian_1955}, which maximises the similarity or in this case minimises the distance cost. For the point detection case, the Hungarian algorithm is thresholded with a radius of $\alpha$, so the algorithm is optimised for number of \textit{TPs} in addition to the computed distance. Typically, a balanced $F_1$ score (e.g., in \cite{sharan_point_2021}) is used to combine the TP, FP, and FN points in a single metric. However, the $F_1$ score is non-monotonic with respect to detections and therefore, we use a \textit{Jaccard} formulation (see Eq.\ref{eq:det_acc}) to compute the detection accuracy, similar to \citep{luiten_hota_2021},

\begin{equation}
detAcc = \frac{|TP|}{|TP|+|FP|+|FN|}.
\label{eq:det_acc}
\end{equation}

It is to be noted that this \textit{Jaccard} score is computed on the set of TPs, FPs, and FNs that are obtained after thresholding. The \textit{Intersection over Union} that is computed as a measure of similarity in an object detection use-case is itself a \textit{Jaccard} metric (Fig. \ref{fig:overview}(Ia)) where this is referred to as a \textit{spatial} \textit{Jaccard} index \citep{luiten_hota_2021}. 

\subsection{Temporal Association}
\label{subsec:temp_assc}
Besides evaluating the detections in every frame, we additionally compute the temporal association as defined for the HOTA metric \citep{luiten_hota_2021} for each \textit{True Positive} (TP) point. For each TP, the \textit{True Positive Association (TPA)} is the number of points in the sequence having the same ground-truth and prediction IDs \citep{luiten_hota_2021}. The points with the same ground-truth ID but a different prediction ID are counted as FNA, and a same prediction ID with a different ground-truth ID is counted as a FPA. The temporal association (\textit{tempAssc}) is then computed as,

\begin{equation}
tempAssc = \frac{1}{|TP|}\sum_{c \in \{TP\}}\frac{|TPA(c)|}{|TPA(c)|+|FPA(c)|+|FNA(c)|}.
\label{eq:ass_acc}
\end{equation}

The formulation of HOTA weights each \textit{TP} by its \textit{tempAssc} value, to combine the detection accuracy and temporal association in a balanced metric, as detailed in \cite{luiten_hota_2021}.

\subsection{Multi-view association}
\label{subsec:mv_assc}

\begin{figure}[t]
\includegraphics[width=\textwidth]{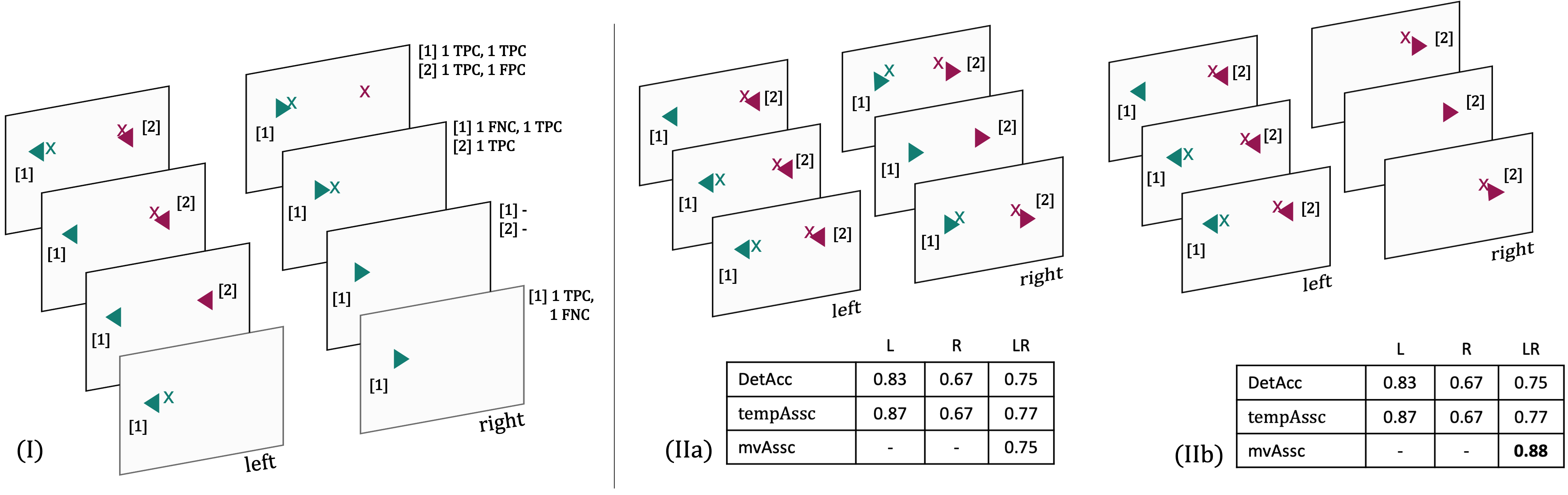}
\caption{An extension to the HOTA metric that incorporates a multi-view setup. (I) An illustration of the different cases of multi-view association when ground-truth spatial association exists (Pt. [1]), and does not exist (Pt. [2]). (IIa,b) An example that illustrates a change in the \textit{mvAssc} without a change in detection (\textit{detAcc}) and temporal association (\textit{tempAssc}) (ground-truth from the $\blacktriangleleft$: left camera view, $\blacktriangleright$: right camera view, \textbf{x}: predictions)}.
\label{fig:example_cases}
\end{figure}

As an extension to the previously proposed HOTA metric \citep{luiten_hota_2021}, we introduce the concept of a matched spatial association between multiple views. 

The multi-view spatial association is defined for a TP in at least one of the views, where it is classified as a True Positive Correspondence (TPC) if the points containing the same ground-truth ID in each of the views are matched with the same prediction ID. If the same ground-truth ID in a corresponding view is matched with a different prediction ID or if there is no prediction found, it is classified as a False Negative Correspondence (FNC). A False Positive Correspondence (FPC) occurs when the same prediction ID in a corresponding view is either matched with another ground-truth ID or if there is no ground-truth found.

The correspondence accuracy can then be defined as the \textit{Jaccard} score averaged for all TP points for each frame in the sequence, computed as

\begin{equation}
mvAssc = \frac{1}{|TP|}\sum_{c \in \{TP\}}\frac{|TPC(c)|}{|TPC(c)|+|FPC(c)|+|FNC(c)|}.
\label{eq:corres_acc}
\end{equation}

Fig. \ref{fig:example_cases} (I) illustrates the possible scenarios for a stereo setup.
For example, point $[1]$ in Fig. \ref{fig:example_cases} (I) is visible in both views. Here, a TPC is defined when a TP is found in both views. An FNC is defined when there are no TP detections found for this $p_{gt}$ in one or both the views. Point $[2]$ in Fig. \ref{fig:example_cases} (I), is present in only one of the views. This is a typical scenario due to occlusions or differences in field of view. Here, a TPC is defined only when a TP detection exists for the $p_{gt}$ in its respective view, and an FPC occurs when the $p_{pred}$ is found in both frames. The formulation of the \textit{mvAssc} similar to the \textit{tempAssc} helps analyse the performance of a model in the context of how well the spatial occlusions are handled. Additionally, separating the spatial from temporal associations, helps measure both the accuracies in a manner that is decomposable into the two different aspects of model performance.

\subsection{Multi-view HOTA}
\label{subsec:mvhota}
We now combine the above mentioned concepts of detection, temporal and spatial associations to formulate the multi-view higher order tracking accuracy: 

\begin{equation}
mvHOTA = \sqrt[3]{detAcc \cdot tempAssc \cdot mvAssc}
\label{eq:mv_hota}
\end{equation}

Each of \textit{tempAssc} and \textit{mvAssc} are averaged over the TP detections, and therefore can be seen as augmenting each TP  with the \textit{tempAssc} and \textit{mvAssc} instead of double counting the errors. Furthermore, \textit{mvHOTA} can be interpreted as the geometric mean of the detection, temporal association, and spatial association, thereby providing equal weighting to each of the factors. Additionally, the matching can be performed to optimise the number of True Positive predictions that contain True Positive Correspondences, in addition to minimising the distance cost and maximising the temporal associations.

\subsection{Occlusion Index (\textit{OI})} 
Besides \textit{mvHOTA}, we propose an 
index to quantify the extent of occlusions in both the temporal and spatial domains. The \textit{OI} is computed as,

\begin{equation}
OI_{v} = 1 - \frac{1}{N} \sum_{f=1}^{N} p^v_{f}. c_f \quad\text{where}\quad  c_f = \frac{1}{M} {\sum_{v=1}^{M} p^f_{v}}
\label{eq:occ_index}
\end{equation}

where $p^f_v$ is used while summing all the views in which a point is found, for a particular frame $f$, where $p^f_v=1$ if the point is found in view $v$ and $0$ otherwise. Similarly, $p^v_f$ is used while summing all the frames in a particular view $v$. 
Essentially, to compute the $OI_v$ for a view $v$, a weighted average is computed for all the frames $f$, where the weighting factor for each point is the fraction of views in which this point exists ($c_f$). 
This index helps in analysing the extent of spatio-temporal occlusions for different objects in the scene.
The temporal occlusion (\textit{tempOI}) and multi-view occlusion (\textit{mvOI}) can be easily obtained from Eqn. \ref{eq:occ_index} by setting $p^f_v=1$ for views $v$, and by setting $p^v_f=1$ for frames $f$ respectively. An example of OI computed for the endoscopic dataset used in this work is shown in Fig. \ref{fig:occ_index}.

\subsection{Toy example}
\begin{figure}[h!]
\centering
\includegraphics[width=\textwidth]{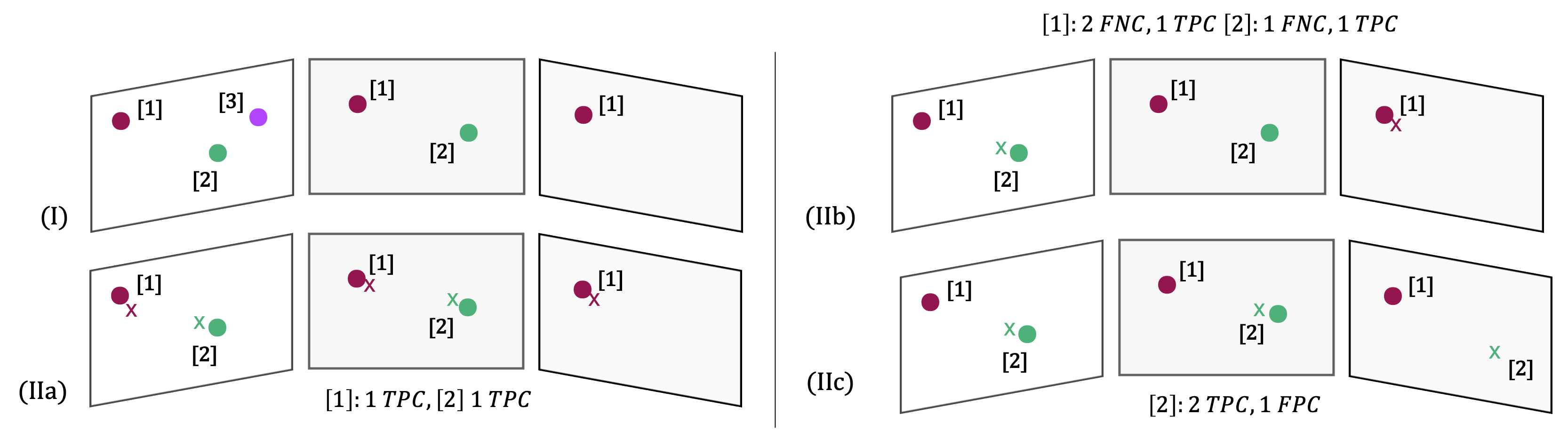}
\caption{An example illustration of a $3$-view setup with ground-truth ($\bullet$) and predicted(\textbf{x}) points. (I) shows how the multi-view spatial association is calculated for each point in different scenarios. (IIa), (IIb), and (IIc) show the calculation of TPC, FPC, and FNC in each case, which is then used to compute the respective \textit{mvAssc}.}
\label{fig:3_view}
\end{figure}

To demonstrate the need for \textit{mvHOTA} in contrast to a trivial multi-view extension of MOT metrics, we construct a toy example to show how \textit{mvHOTA} can accommodate the spatial association from different views while related MOT metrics remain unchanged. Fig. \ref{fig:example_cases} (II) represents an example case in point detection with two ground-truth IDs. Fig. \ref{fig:example_cases} (IIa) contains two ground-truth points and their respective predictions. In Fig. \ref{fig:example_cases} (IIb), we now remove one of the ground-truth IDs and the associated predictions. This changes the spatial association without affecting the detection and temporal association. Here, \textit{mvHOTA} is able to accommodate this change, while computing the MOT metrics for each view (see Tab. \ref{tab:toy_table_data}) does not account for the spatial association. A further example of \textit{mvHOTA} applied to more than two views is shown in Figure \ref{fig:3_view}. 

\begin{table}[htb]
\tbl{Comparison of MOT metrics on the two cases of the toy example as shown in Fig. \ref{fig:example_cases} (II). Change in \textit{mvHOTA} is highlighted in bold.}
{\begin{tabular}{l|ccccc}
Case   &  MOTA &  IDF1 &  F1   & HOTA &  mvHOTA         \\ \midrule
(II)a. &  0.67 &  0.76 &  0.85 & 0.76 & 0.75            \\
(II)b. &  0.67 &  0.76 &  0.85 & 0.76 & \bf{0.88}       \\
\end{tabular}}
\label{tab:toy_table_data}
\end{table}

\subsection{Properties}
For a single-view, single-instance tracking scenario, \textit{mvHOTA} simplifies into $|TP|/(|TP|+|FP|+|FN|)$. \cite{luiten_hota_2021} showed that HOTA is both monotonic and differentiable into the different error types, as proposed by \cite{leichter_10.1109/TPAMI.2013.70}. \textit{mvHOTA} follows a similar formulation while including the spatial associations between the multiple views. Furthermore, the interpretation as the geometric mean ensures the metric is balanced between the different aspects of detection, temporal and multi-view spatial associations. The ability to differentiate the metric into different error types \citep{leichter_10.1109/TPAMI.2013.70} and analyse their aspects is especially crucial for benchmarking, ranking, and interpreting the performance of the different methods. 
Evaluating a metric with respect to different aspects of performance not only helps in comparing methods, but also in aligning this comparison towards an application that favours specific aspects of performance. In this regard, \cite{leichter_10.1109/TPAMI.2013.70} define $5$ basic error types for MOT tasks, namely \textit{False Negatives}, \textit{False Positives}, \textit{Fragmentation}, \textit{Mergers}, and \textit{Deviations}. \cite{luiten_hota_2021} showed how for the tracking task HOTA decomposes into the five error types with an equivalent computation of detection recall, detection precision, association recall, association precision, and localisation. \textit{mvHOTA} that is built upon this work can be similarly analysed with respect to the basic error types.

% -------------- EXPERIMENTS ------------------------------------------------
\section{Experiments and Results}
\label{sec:experiments}

In this section we demonstrate the application of the proposed metric to evaluate the tracking of points for a stereo-endoscopic use-case in mitral valve repair \citep{engelhardt_deep_2021}.

\subsection{Dataset}

\begin{figure}[h!]
\centering
\includegraphics[width=1\textwidth]{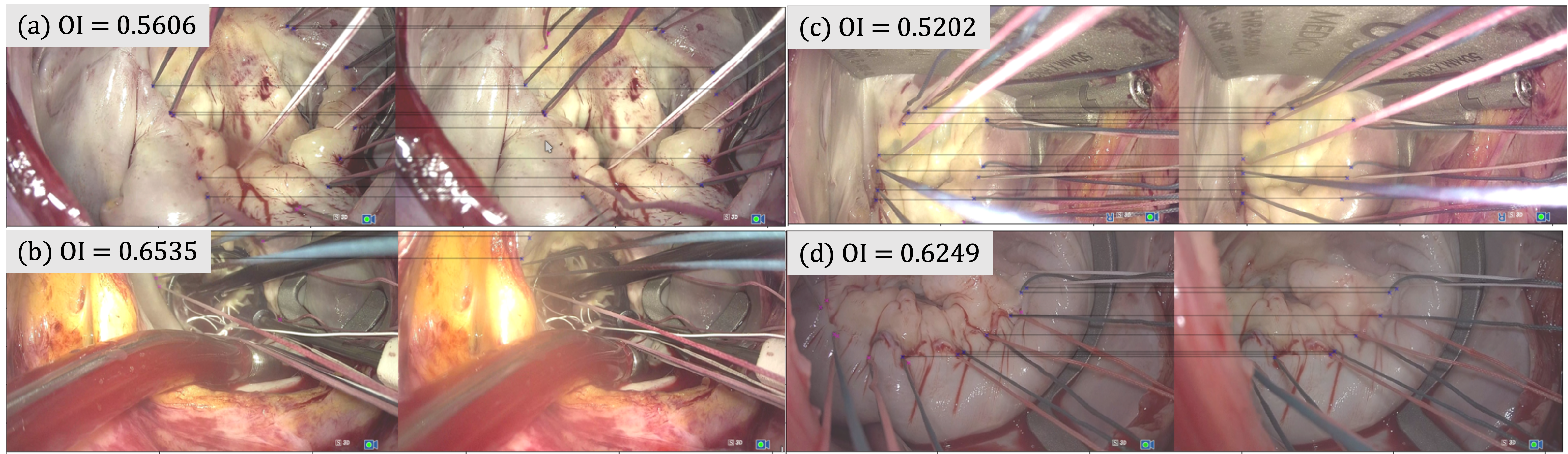}
\caption{Sample images from Folds $1-4$ of the mitral valve training dataset with the corresponding left and right views of the stereo-endoscopic setup are shown in (a-d) respectively, with the spatial associations drawn for the labelled suture points. A point can be spatially occluded due to the presence of sutures, instruments, or occluding tissue in the scene. The respective \textit{Occlusion Indices (OIs)} of the surgeries are additionally shown, that indicate the average extent of occlusions in both the temporal and spatial domains.}
\label{fig:occ_index}
\end{figure}

The mitral valve dataset \citep{engelhardt_deep_2021,sharan_point_2021} comprises images from a stereo-endoscope captured during minimally invasive mitral valve repair surgery. 
The dataset contains surgeries captured under varying camera angles and illumination. Moreover, objects moving in an out of the scene (surgical instruments, suction pump, ring sizer, etc.) occlude points of interest (entry and exit points of sutures) in both the temporal and spatial domains.
A dataset of $4$ surgeries was used for training.

\begin{figure}[h!]
\centering
\includegraphics[width=1\textwidth]{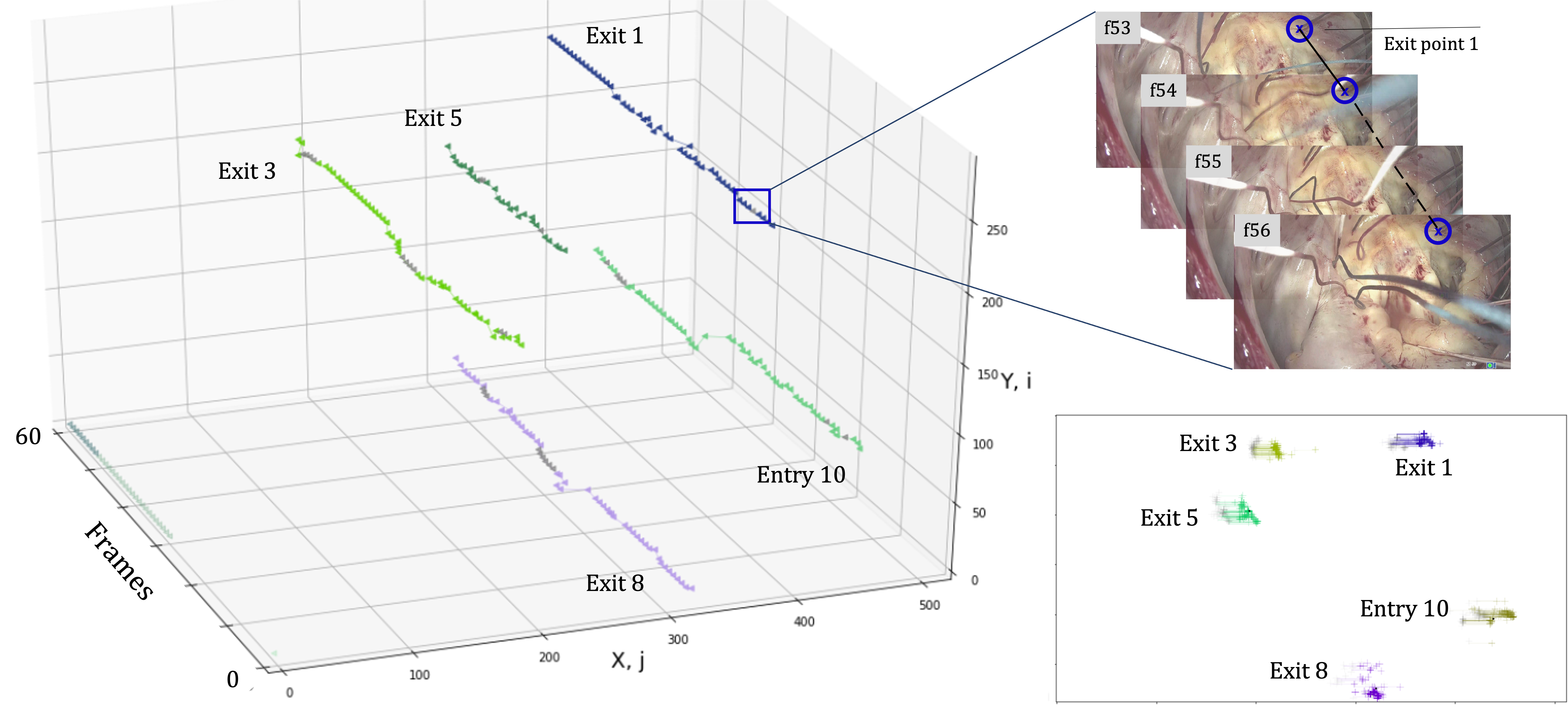}
\caption{Sample trajectories of sample entry and exit suture points from Surgery $1$ of the training dataset, for the first $60$ frames of the temporal sequence. The points plotted in gray represent temporal occlusions, where an interpolated value is plotted in the place of the true suture point location. Additionally, a closeup of Exit point $1$ is shown for frames $53-56$ where a temporal occlusion due to the presence of white sutures can be seen in frame $55$. Additionally, a cluster projection on the $x$-$y$ plane of the whole sequence of the suture points are presented on the lower right, to illustrate the variance in the movement across the sequence. Furthermore, the disparities with the corresponding view of the stereo-setup are drawn for each of the points.}
\label{fig:examples}
\end{figure}

A sample from each surgery of the training dataset with the left and right views of the multi-view setup is shown in Fig. \ref{fig:occ_index} with the spatial associations highlighted. The \textit{Occlusion index (OI)} is provided for the respective datasets, which indicate the average of extent of occlusion in both the temporal and spatial domains.
Furthermore in Fig. \ref{fig:examples}, the sample trajectories of entry and exit suture points across a temporal sequence is illustrated for Surgery $1$ of the training dataset, together with a closer look at one sample of a suture point. Additionally, Fig. \ref{fig:examples} also shows the variance of the projected suture clusters for the whole sequence along with the disparity shifts between the corresponding views for the respective suture points.

\begin{figure}[h!]
\centering
\includegraphics[width=0.9\textwidth]{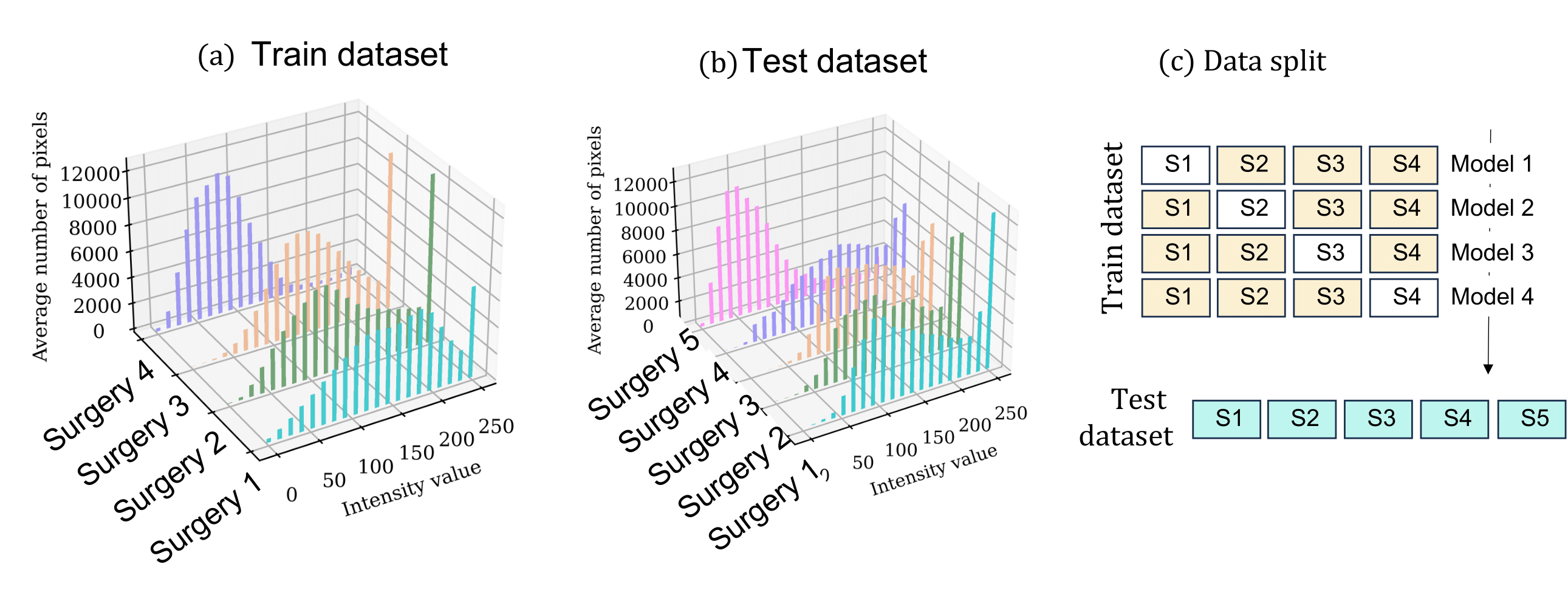}
\caption{(a, b) show the intensity distribution of the different surgeries in the train and test datasets respectively. (c) shows the data split used for the experiments.} 
\label{fig:data_split}
\end{figure}

With this training dataset, the models were trained with leave-one out cross-validation, yielding $4$ different models for analysis. The trained models were tested on an external test dataset comprising of $5$ challenging surgeries that contain occlusions in both the spatial and temporal domains. 
Fig. \ref{fig:data_split}(a, b) illustrate the intensity distributions of the training and test dataset respectively. Fig. \ref{fig:data_split}(c) provides an overview of the data split used in the experiments in this work.

The multi-point detection task is to detect the entry and exit points of sutures, which are stitched. The suture points do not occur at anatomically unique locations and the number of suture points vary in each image temporally and between the views. The endoscopic frames are publicly available as part of the AdaptOR challenge \citep{engelhardt_deep_2021}. Other works on this data set have primarily focused on the detection task and treated the stereo-information as two mono instances. The authors have formulated it as a multi-instance heatmap regression problem \citep{sharan_point_2021,stern_10.1007/978-3-658-33198-6_7}. 
Evaluation was performed by computing a balanced $F_1$ score. A threshold of $6\mathrm{px}$ was set as the similarity threshold, as it roughly corresponds to the thickness of the suture in this image resolution. However, a frame level metric such as the $F_1$ score does not indicate the performance of the method in handling temporal and spatial occlusions. Therefore it is important to compute the respective associations in the temporal and spatial domains to better ascertain the performance of the method. 

\begin{figure}[h!]
\centering
\includegraphics[width=0.9\textwidth]{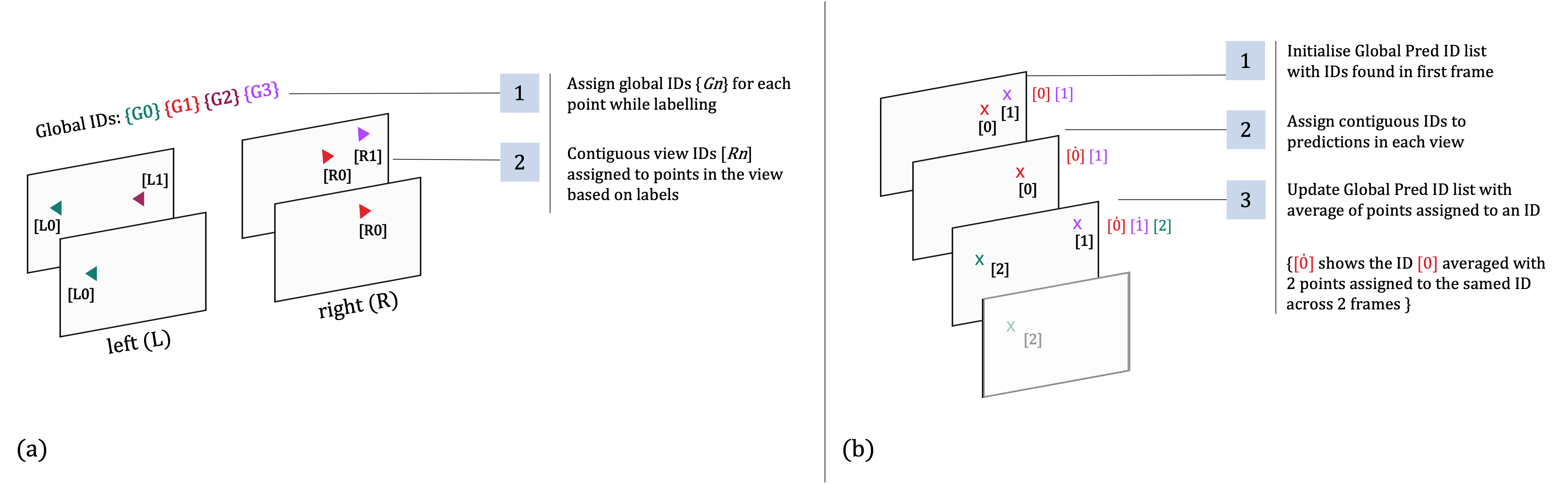}
\caption{The use of mvHOTA requires temporally matched IDs assigned to both the ground-truth ($\blacktriangleleft$: left camera view, $\blacktriangleright$: right camera view), and predicted (\textbf{x}) points. This figure shows how (a) global ground-truth IDs are created based on labels, but are mapped to locally contiguous IDs assigned to each view. (b) Predicted points are temporally matched to a global list for each view, which is averaged after every frame is matched.}
\label{fig:id_assignment}
\end{figure}

\subsection{Data preparation}
\label{subsec:id}

In order to compute the multi-view higher order tracking accuracy for the suture detection method on the mitral valve dataset, we show how we can prepare the dataset. Firstly, the computation of temporal associations requires contiguous ID assignment in each view. This is achieved by mapping the IDs of each view to a global ID list that is common to both views (see Fig. \ref{fig:id_assignment} (a)). Secondly, we require unique contiguous IDs for a sequence of images for the predicted points. This is achieved by temporally matching the predicted points with the Hungarian matching algorithm (see Fig. \ref{fig:id_assignment} (b)).

\subsection{Results}
\label{subsec:results}

\begin{table}[htb]
\tbl{Comparison of MOT metrics for each fold (surgery) of the mitral valve dataset. The MOTA, IDF1, F1, and HOTA metrics are averaged for each view.}
{\begin{tabular}{l|cccccccc} 
Metric     & MOTA     & IDF1    & F1      & detAcc   & tempAssc   & HOTA    & mvAssc   & mvHOTA  \\ \midrule
Model 1    & -0.0256  & 0.1559  & 0.3892  & 0.2629   & 0.4109     & 0.3241  & 0.7363   & 0.4172  \\ 
Model 2    & 0.0798   & 0.1815  & 0.3902  & 0.2619   & 0.4040     & 0.3141  & 0.7283   & 0.4104  \\ 
Model 3    & 0.0661   & 0.1687  & 0.3815  & 0.2498   & 0.4686     & 0.3398  & 0.7241   & 0.4339  \\ 
Model 4    & 0.0786   & 0.1572  & 0.3425  & 0.2264   & 0.4098     & 0.3034  & 0.7193   & 0.3868  \\ \midrule
Mean       & 0.0497   & 0.1658  & 0.3759  & 0.2503   & 0.4233     & 0.3204  & 0.7270   & 0.4121  \\ 
\end{tabular} }
\label{tab:endo_table_data}
\end{table}

Table \ref{tab:endo_table_data} presents a comparison of various MOT metrics computed for each surgery of the mitral valve test dataset for the suture tracking task. 
The various models are trained on different splits of training dataset using leave-one out cross-validation and tested on all the surgeries of the test dataset (c.f. Fig. \ref{fig:data_split}(c)). Besides, the temporal IDs for the predictions were assigned in a post-processing step as described in Section \ref{subsec:id}.
It can be seen that although the HOTA and \textit{mvHOTA} provide similar comparative trends of the different folds, the \textit{mvHOTA} metric provides additional specific information for analysis of the spatial associations in a dataset. 

\begin{figure}[h!]
\centering
\includegraphics[width=1\textwidth]{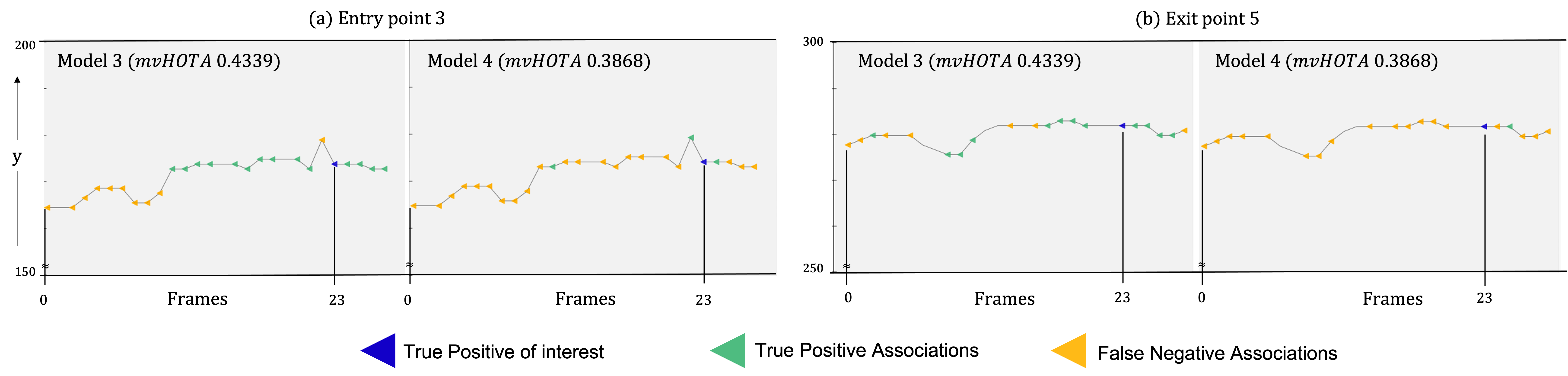}
\caption{The association trajectory over time over the first $30$ frames, for (a) Entry point $3$, and (b) Exit point $5$ from Surgery $1$ of the test dataset, for the True Positive found in Frame $23$. The trajectories of the suture point are projected on the $x$-$t$ plane, and the True Positive Associations (TPA) and the False Negative Associations (FNA) are shown. Model $3$ with \textit{mvHOTA} $=0.4339$ outperforms Model $4$ with \textit{mvHOTA} $=0.3868$.}
\label{fig:comparison_1}
\end{figure}

\begin{figure}[h!]
\centering
\includegraphics[width=1\textwidth]{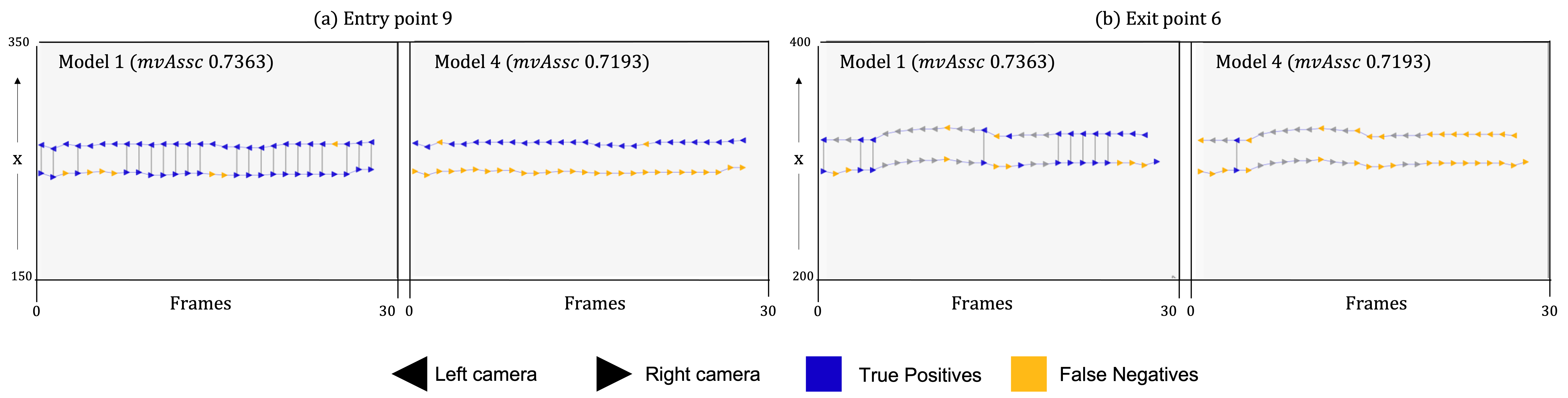}
\caption{
The trajectory over time of the first $30$ frames, for (a) Entry point $9$, and (b) Exit point $6$ from Surgery $3$ of the test dataset. The trajectories of the suture point are projected on the $y$-$t$ plane. The True Positive Correspondences (TPC) are connected with a line, whereas the False Negative Correspondences (FNC) are not. Model $1$ with \textit{mvAssc} $=0.7363$ outperforms Model $4$ with \textit{mvAssc} $=0.7193$.
}
\label{fig:comparison_2}
\end{figure}

Model $3$ has the highest \textit{mvHOTA} score of $0.4339$, and the highest \textit{HOTA} score of $0.3398$. However, it does not have the highest detection performance, which is reflected in a \textit{detAcc} value of $0.2498$ (compared to $0.2629$ of Model $1$ c.f. Table \ref{tab:endo_table_data}), which is also reflected in the frame level $F_1$ score of $0.3815$ (compared to $F_1=0.3902$ for Model $2$ c.f. Table \ref{tab:endo_table_data}). In contrast, Model $4$ has the lowest \textit{mvHOTA} score of $0.3868$ ($-0.0471$ of Model $3$) and the lowest \textit{HOTA} score of $0.3034$ ($-0.0364$ of Model $3$). Accordingly, Model $3$ has a \textit{tempAssc} of $0.4686$, compared to Model $4$ with the lowest \textit{tempAssc} of $0.4098$ ($+0.0588$ c.f. Table \ref{tab:endo_table_data}). 
Two such examples, are illustrated in Fig. \ref{fig:comparison_1}(a) and (b), which indicate the difference in the performance of Model $3$ and $4$, with respect to temporal associations on the same data, here namely Entry point $3$, and Exit point $5$ respectively. It can be seen that Model $3$ has more temporal associations for a given TP in comparison with Model $4$.
Similarly, the \textit{mvAssc} of Model $1$ is the highest ($0.7363$), and that of Model $4$ the lowest ($0.7193$, $-0.0170$ c.f. Table \ref{tab:endo_table_data}). 
Fig. \ref{fig:comparison_2} illustrates this difference, for Entry point $9$ (Fig. \ref{fig:comparison_2}(a)), and Exit point $6$ (Fig. \ref{fig:comparison_2}(b)), which show that Model $1$ is able to predict the sutures in both the views more consistently, and as a result has a higher aggregate \textit{mvAssc} compared to Model $4$ which has a large number of FNCs
Additionally, it can be seen that Model $1$ has a negative MOTA score ($-0.0256$), which means the combined value of FPs, FNs, and \textit{Identity Switches (IDSW)} as defined by \citep{bernardin_evaluating_2008} are in this case, more than the number of ground truth detections \citep{bernardin_evaluating_2008}.

% -------------- DISCUSSION------------------------------------------------
\section{Discussion and Conclusion}
\label{sec:discussion}

The proper choice of metrics for a particular task is non-trivial, especially in the case of multi-point detection due to objects moving in and out of the scene, and endoscopic artefacts, leading to spatial and temporal occlusions. Incorporating these associations provides a more complete picture in evaluating the performance of multi-point detection methods. 
While a range of metrics have been previously proposed to evaluate a MOT task \citep{bernardin_evaluating_2008,vedaldi_tao_2020,ristani_10.1007/978-3-319-48881-3_2}, in this work we propose a metric to track temporal and multi-view spatial associations in multi-point tracking.
Beyond the use case of mitral valve repair, different endoscopic datasets, be it laparoscopy \citep{Bodenstedt2018ComparativeEO} or heart surgery \citep{sharan_point_2021}, contain varying amounts of temporal or spatial occlusions depending on the moving or static nature of the camera and the scene. The \textit{mvHOTA} metric is especially useful in this case, as it enables analysis and benchmarking of different aspects of a model performance, while at the same time can be decomposed into different aspects of tracking for a multi-camera setup.

Alternatively, computing HOTA \citep{luiten_hota_2021} for each view and averaging them, does not embed the multi-view spatial associations (see Fig. \ref{fig:example_cases} (IIb)). \cite{luiten_hota_2021} suggest a multi-camera extension by stacking the frames of the corresponding view together with the temporal stack to evaluate the total associations for the sequence. However, this precludes an analysis of the temporal and spatial associations in a separate manner while combining them in a single metric. 

However, there exist some limitations of \textit{mvHOTA}. \cite{luiten_hota_2021} compute the localisation accuracy for a particular threshold $\alpha$, to measure the extent of spatial alignment between a ground-truth and predicted point for the object detection case, based on the \textit{IoU} metric. A similar metric can also be computed for this use-case as the distance cost of the points that lie within the threshold $\alpha$. However, it is not shown in this work, since the goal is to quantify the detection and tracking performance for a particular threshold.
The distance cost between the ground-truth and predicted points is however used to optimise the matching.
A future work is to generalise this method to object detection use-cases where the detection and tracking performance across a multi-view setup can be quantified.

\section*{Funding}
This work was supported in part by Informatics for Life funded by the Klaus Tschira Foundation and the German Research Foundation DFG Project 398787259, DE 2131/2-1 and EN 1197/2-1.).

\section*{Ethics Declarations}
The study was approved by the Local Ethics Commitee from Heidelberg University Hospital. Registration numbers
are S-658/2016 (12.09.2017) and S-777/2019 (20.11.2019). Informed written consent to take part in the research was obtained prior to the commencement of the study.

\section*{Disclosure statement.}
The authors report there are no competing interests to declare. 

\section*{Biographical Note}
\begin{itemize}
    \item \textbf{Lalith Sharan} is currently a doctoral student in the group Artificial Intelligence in Cardiovascular Medicine at the Heidelberg University Hospital. His research interests include computer-assisted surgeries, depth estimation, object and landmark detection.
    
    \item \textbf{Halvar Kelm} received his B.Sc. in Medical Informatics from the Mannheim University of Applied Sciences. He is currently pursuing his Master's degree in Informatics at the University of Bremen. His research interests include deep learning, computer vision, and federated learning for medicine.  

    \item \textbf{Gabriele Romano} is a resident surgeon at the Department of Cardiac Surgery of the Heidelberg University Hospital, and a co-operation partner with the group Artificial Intelligence in Cardiovascular Medicine. His research interests include minimally invasive mitral valve surgery, in particular surgical simulation for surgical training and planning. 
    
    \item \textbf{Prof. Dr. Matthias Karck} is the Medical Director of the Department of Cardiac Surgery at the Heidelberg University Hospital. He is an expert in mitral valve repair, aortic surgery, and aortic root reconstruction.
    
    \item \textbf{Raffaele De Simone} is an expert Cardiac surgeon and a Cardiologist at the University Hospital Heidelberg. His research interests include mitral valve regurgitation jets and assistant systems for reconstructive heart valve surgery.
    
    \item \textbf{Sandy Engelhardt} is an Assistant Professor at the University Hospital Heidelberg, and leads the group Artificial Intelligence in Cardiovascular Medicine. Her research interests include image analysis and visualisation for cardio-vascular diseases.
\end{itemize}

\section*{Data availability statement}
The data that support the findings of this study are previously released publicly as part of the AdaptOR Challenge 2021 at \url{https://adaptor2021.github.io/} organised in the Medical Image Computing and Computer Assisted Interventions Conference (MICCAI) 2021, and is available from the challenge organisers upon reasonable request.

\bibliographystyle{chicago}
\bibliography{references.bib}
\end{document}